\def\year{2021}\relax
\documentclass[letterpaper]{article} 
\usepackage{aaai21}  
\usepackage{times}  
\usepackage{helvet} 
\usepackage{courier}  
\usepackage[hyphens]{url}  
\usepackage{graphicx} 
\usepackage{natbib}  
\usepackage{caption} 
\usepackage[utf8]{inputenc} 
\usepackage[T1]{fontenc}    
\usepackage{url}            
\usepackage{booktabs}       
\usepackage{amsfonts}       
\usepackage{nicefrac}       
\usepackage{microtype}      
\usepackage{makecell}
\usepackage{multirow}
\usepackage{algorithmic}
\usepackage{amsmath}
\usepackage{amssymb}
\usepackage{graphicx}
\usepackage{wrapfig}
\usepackage{lipsum}
\usepackage[ruled]{algorithm2e}
\usepackage[pagebackref=true,breaklinks=true,letterpaper=true,colorlinks,bookmarks=false]{hyperref}

\usepackage{pifont}

\usepackage[switch]{lineno}  %

\title{Partial Is Better Than All: \\Revisiting Fine-tuning Strategy for Few-shot Learning}
\author{
    Zhiqiang Shen\textsuperscript{\rm 1$^{\ast}$}, Zechun Liu\textsuperscript{\rm 1,2\thanks{indicates equal contribution.}}, Jie Qin\textsuperscript{\rm 3}, Marios Savvides\textsuperscript{\rm 1} and Kwang-Ting Cheng\textsuperscript{\rm 2}
}
\affiliations{

    \textsuperscript{\rm 1}Carnegie Mellon University \textsuperscript{\rm 2}Hong Kong University of Science and Technology \\ \textsuperscript{\rm 3}Inception Institute of Artificial Intelligence\\

    \{zhiqians,marioss\}@andrew.cmu.edu;zliubq@connect.ust.hk;qinjiebuaa@gmail.com;timcheng@ust.hk

}

\begin{document}


\maketitle

\begin{abstract}
The goal of few-shot learning is to learn a classifier that can recognize unseen classes from limited support data with labels. A common practice for this task is to train a model on the base set first and then transfer to novel classes through fine-tuning\footnote{Here fine-tuning procedure is defined as transferring knowledge from base to novel data, i.e. learning to transfer in few-shot scenario.} or meta-learning. However, as the base classes have no overlap to the novel set, simply transferring whole knowledge from base data is not an optimal solution since some knowledge in the base model may be biased or even harmful to the novel class. In this paper, we propose to transfer partial knowledge by freezing or fine-tuning particular layer(s) in the base model. Specifically, layers will be imposed different learning rates if they are chosen to be fine-tuned, to control the extent of preserved transferability. To determine which layers to be recast and what values of learning rates for them, we introduce an evolutionary search based method that is efficient to simultaneously locate the target layers and determine their individual learning rates. We conduct extensive experiments on CUB and {\em mini}-ImageNet to demonstrate the effectiveness of our proposed method. It achieves the state-of-the-art performance on both meta-learning and non-meta based frameworks. Furthermore, we extend our method to the conventional {\em pre-training + fine-tuning} paradigm and obtain consistent improvement.
\end{abstract}

\begin{figure}[t]
    \centering
    \includegraphics[width=0.87\linewidth]{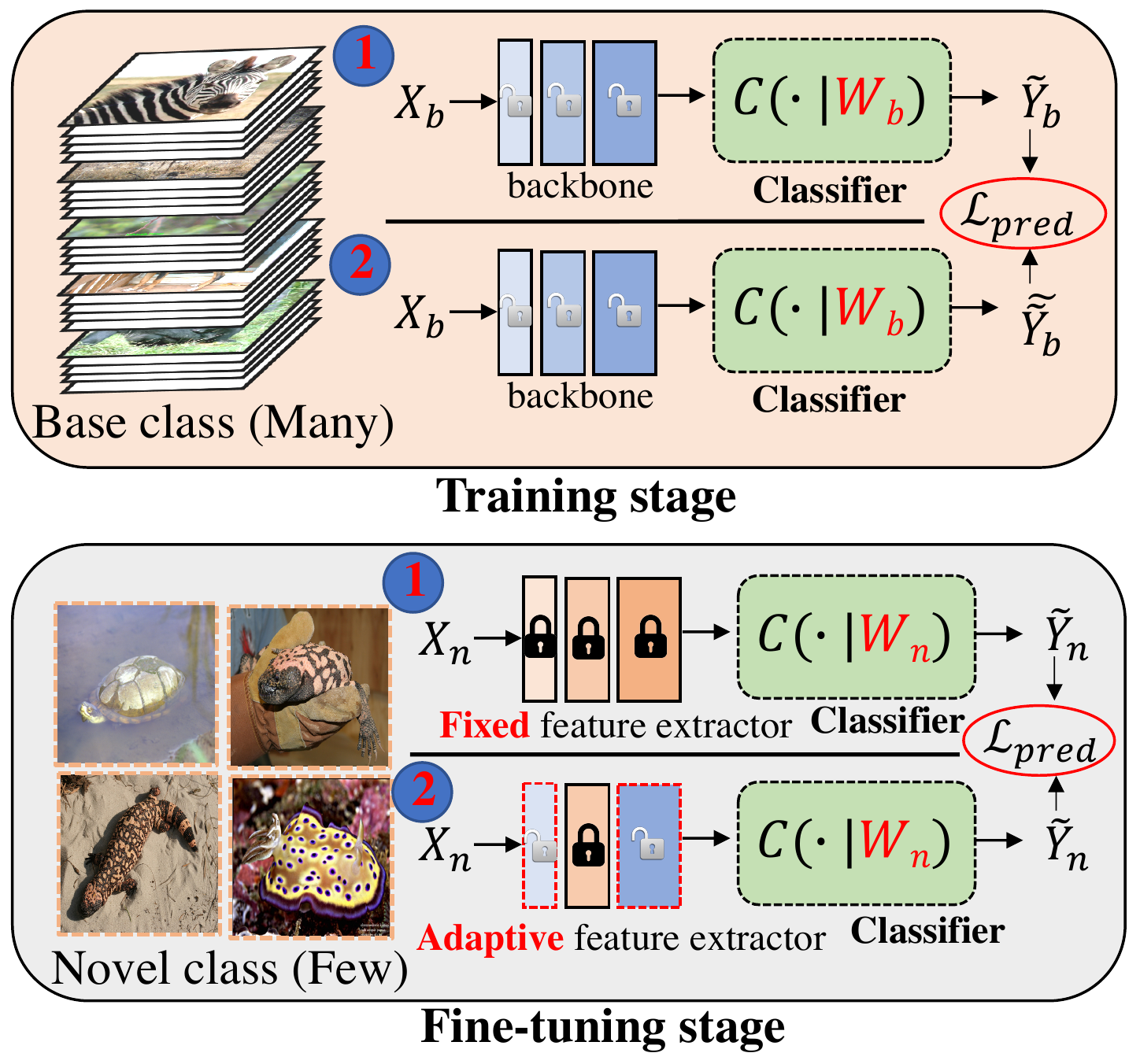}
    \vspace{-0.12in}
    \caption{Illustration of the conventional procedure of pre-training and fine-tuning for few-shot learning. \ding{172} represents the standard transfer learning procedure which uses the pre-trained model as a feature extractor and the parameters are fixed during fine-tuning. \ding{173} is our proposed partial transfer strategy which can fine-tune the model trained on base data with the few novel class data. Fine-tuning with different learning rates on different layers can optimize the feature extractor to better fit the novel class and prevent the model from overfitting on it, since the novel data has limited samples.}
    \label{fig:comparison}
    \vspace{-0.1in}
\end{figure}

\section{1. Introduction}
Deep neural networks have shown enormous potential on understanding natural images~\cite{krizhevsky2012imagenet,szegedy2015going,simonyan2014very,he2016deep,huang2017densely} in the recent years. The learning ability of deep neural networks increases significantly with more labeled training data. However, annotating such data is expensive, time-consuming and laborious. Furthermore, some classes (e.g., in medical images) are naturally rare and hard to collect. The conventional training approaches for deep neural networks often fail to obtain good performance when the training data is insufficient. Considering that humans can easily learn from very few examples and even generalize to many different new images, it will be greatly helpful if the network can also learn to generalize to new classes with only a few labeled samples from unseen classes. Previous studies in this direction (i.e., few-shot learning) can be mainly divided into two categories. One is the meta-learning based methods~\cite{snell2017prototypical,vinyals2016matching,finn2017model} that model the few-shot learning process with samples belonging to the base classes, and optimize the model for the target novel classes. The other is the plain solution (non-meta and also called baseline method~\cite{chen2018a}) that trains feature extractor from abundant base class then directly predicts the weights of the classifier for the novel ones.

As the number of images in the support set of novel classes are extremely limited, directly training models from scratch on the support set is unstable and tends to be overfitting. Even utilizing the pre-trained parameters on base classes and fine-tuning all layers on the support set will still lead to poor performance due to the small proportion of target training data. A common practice utilized by either meta-based or simple baseline methods relies heavily on the pre-trained knowledge with the sufficient base classes, and then transfer the representation by freezing the backbone parameters and solely fine-tuning the last fully-connected layer or directly extracting features for distance computation on the support data, to prevent overfitting and improve generalization. However, as the base classes have no overlap with the novel ones, meaning that the representation and distribution required to recognize images are quite different between them, completely freezing the backbone network and simply transferring the whole knowledge will suffer from this discrepant domain issue, though currently the domain difference is not huge in the existing few-shot learning datasets. 

To fundamentally overcome the aforementioned drawback, in this work, we propose to utilize a flexible way to transfer knowledge from base to novel classes. We introduce a partial transfer paradigm for the few-shot classification task, as shown in Figure~\ref{fig:comparison}. In our framework, we first pre-train a model on the base classes as previous methods did. Then, instead of transferring the learned representation by freezing the whole backbone network, we develop an efficient evolutionary search method to automatically determine which layer/layers need to be frozen and which will be fine-tuned on the support set (on novel class). During searching, the validation data will be commandeered as the ground-truth to monitor the performance of the searched strategy. Intuitively, our strategy can achieve a better trade-off of using knowledge from base and support data than previous approaches, meanwhile, our strategy can avoid incorporating biased or harmful knowledge from base classes into novel classes. Moreover, our method is orthogonal to meta-learning or non meta-based solutions, and thus can be seamlessly integrated with them. We perform extensive experiments on CUB200-2011 and {\em mini}-ImageNet datasets. Our results empirically show that the proposed method can favorably improve both of these two types of solutions. We further extend our method to the traditional {\em pre-training + fine-tuning} paradigm from ImageNet to CUB200-2011 and achieve consistent improvement, demonstrating the effectiveness and excellent expansibility of our proposed method. 

In summary, our contributions are three-fold:

\ \ $\bullet$ We present Partial Transfer (P-Transfer) for the few-shot classification, a framework that enables to search transfer strategies on backbone for flexible fine-tuning. Intuitively, the conventional fixed transferring is a special case of our propose strategy when all layers are frozen. Also, to our best knowledge, this is the pioneer attempt that can achieve partial transfer with different learning rates on this challenging task.

\ \ $\bullet$ We introduce a layer-wise search space for fine-tuning from base classes to novel. It helps the searched transfer strategy obtain inspiring accuracies under limited searching complexity. For example, using one V100 GPU, our search algorithm only takes $\sim$6 hours with Conv6 backbone and one day with ResNet-12 backbone on average.

\ \ $\bullet$ Our resulting network, the P-Transfer model, outperforms the complete transfer and the hand-crafted transfer strategies by a remarkable margin. As the two baseline transfer strategies belong to our search space, thus ideally the better performance is guaranteed by our searching method. With the assistance of designing search space, we show the effectiveness of P-Transfer in different few-shot learning methods on various datasets. The searched strategy has consistently better performance and meaningful structural patterns.

\begin{figure*}[t]
    \centering
    \includegraphics[width=0.9\linewidth]{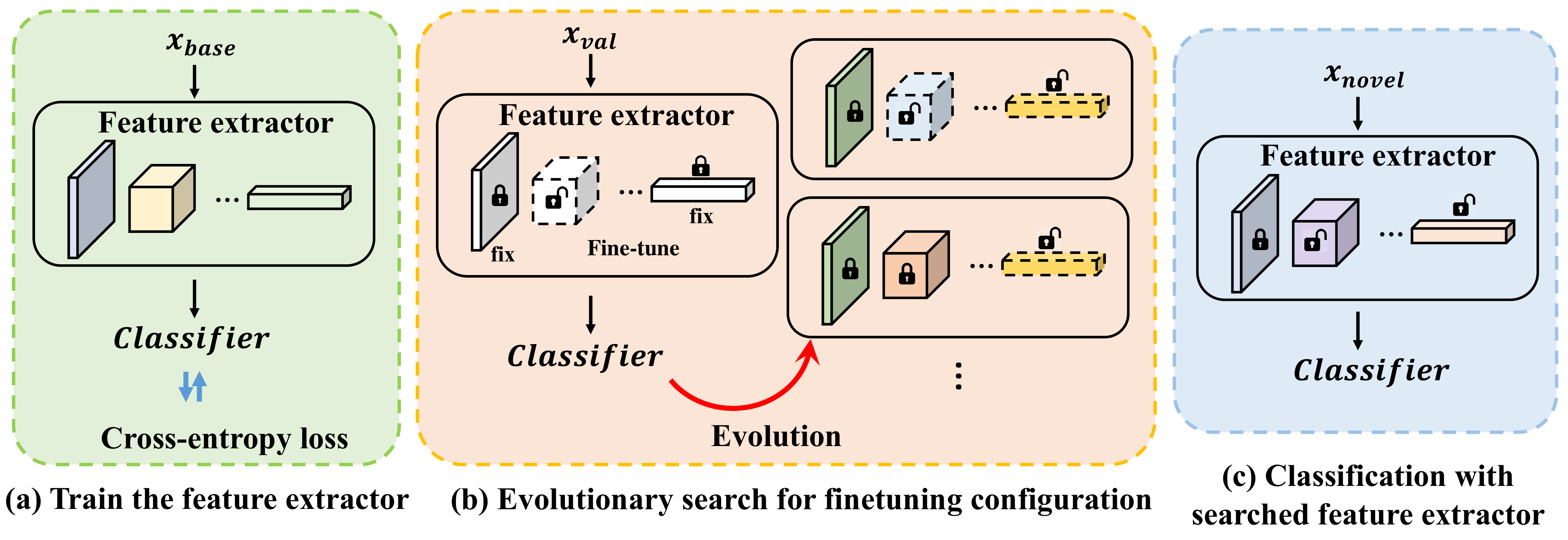}
    \vspace{-0.15in}
    \caption{Our overall framework overview consists of three stages: (a) train a feature extractor from scratch on the base dataset; (b) apply evolutionary search to explore optimal combination of layers that requires fine-tuning on the validation dataset. Note that the blocks with dashed lines denote the fine-tuning layers in a specific evolution iteration; (c) use the best fine-tuning scheme discovered by the evolutionary search to fine-tune the selected layers on the support set of the novel dataset and inferring the final accuracy on the query set.}
    \label{fig:overall_framework}
    \vspace{-0.18in}
\end{figure*}

\section{2. Background}

Few-shot learning is defined as given abundant labeled training examples in the base classes, the trained network can be generalized to classify the new classes with few labeled samples. Recently, few-shot learning, enabled by transferring knowledge from the base to novel data, has been increasingly important. Existing few-shot learning methods can mainly be categorized into meta-learning based methods and non-meta learning methods. Here we also review the searching based methods for few-shot learning in this section. 

\noindent{\textbf{Meta-based few-shot learning.}}
To tackle the data deficiency in few-shot learning, previous studies adopt meta-learning to learn the model or optimizer that can fast update the weights for adapting to the unseen tasks~\cite{thrun2012learning, andrychowicz2016learning,ye2020few,tian2020rethinking,kimmodel}. For example, MetaNetwork~\cite{munkhdalai2017meta} learned a meta-level knowledge for rapid generalization. Ravi and Larochelle~\cite{ravi2016optimization} proposed to use the LSTM-based meta-learner model to learn the optimization algorithm. MAML~\cite{finn2017model,antoniou2018train} simplified the aforementioned MetaNetwork by only learning the initial learner parameters to achieve rapid adaptation w.r.t. those initial parameters and high generalizability to the new tasks. 
Furthermore, meta-learning methods are utilized to learn the similarity between two images. MatchingNet~\cite{vinyals2016matching} proposed to map a small labeled support set to its label, and determine the class of an instance in the query set by finding its nearest labeled example. ProtoNet~\cite{snell2017prototypical} further utilized class-wise mean and the Euclidean distance to generalize the MatchingNet from one-shot learning to few-shot learning. RelationNet~\cite{sung2018learning} use CNN-based relation modules and Few-shot GNN~\cite{garcia2017few} employed graph neural networks to learn useful metrics.

\noindent{\textbf{Non-meta few-shot learning.}}
Besides those meta-learning based methods, there are non-meta methods which utilize cosine similarity to predict the novel class classifier with weight generator~\cite{gidaris2018dynamic}, directly set the weights based on the embedding layer’s activations~\cite{qi2018low} or use dense representations from image regions to calculate the distances~\cite{zhang2020deepemd}. Chen et al.~\cite{chen2018a} proposed to reduce intra-class variation along with the confine similarity and achieves competitive performance. 
Both the meta and non-meta methods used the fixed feature extractor trained from the based classes, which can hardly take the domain discrepancy between the base and novel classes into consideration. Instead of learning more advanced optimizers or classification metrics, we tackle the few-shot problem by discovering an meta knowledge transfer scheme through evolutionary search, which is compatible with both meta and non-meta methods.

\noindent{\textbf{Neural architecture search for few-shot learning.}}
Evolutionary algorithm has been adopted in the neural architecture search (NAS) to obtain the optimal neural architecture~\cite{miikkulainen2019evolving,real2017large,xie2017genetic,liu2017hierarchical,real2019regularized}. Evolutionary-based NAS evolves within a given architecture search space and updates a population of genes (i.e., the operation choices in an architecture) to select the top gene for the final model. Recent study~\cite{elsken2020meta} proposed to integrate NAS algorithm with gradient based meta-learning for few-shot learning task. Different from neural architecture search, our proposed P-Transfer utilizes evolutionary algorithm to seek the optimal fine-tuning scheme, instead of the network architecture. Meta-SGD~\cite{li2017meta} and MAML++~\cite{antoniou2018train} can also learn diverse learning rates for each layer in the networks, but they were mainly designed for MAML-like methods and only suitable for the meta-based scenarios. In contrast, our proposed method can completely turnoff the learning rate to zero and fix the weights in a layer, which is a more general design for the few-shot learning task.

\begin{figure*}[t]
    \centering
    \includegraphics[width=0.73\linewidth]{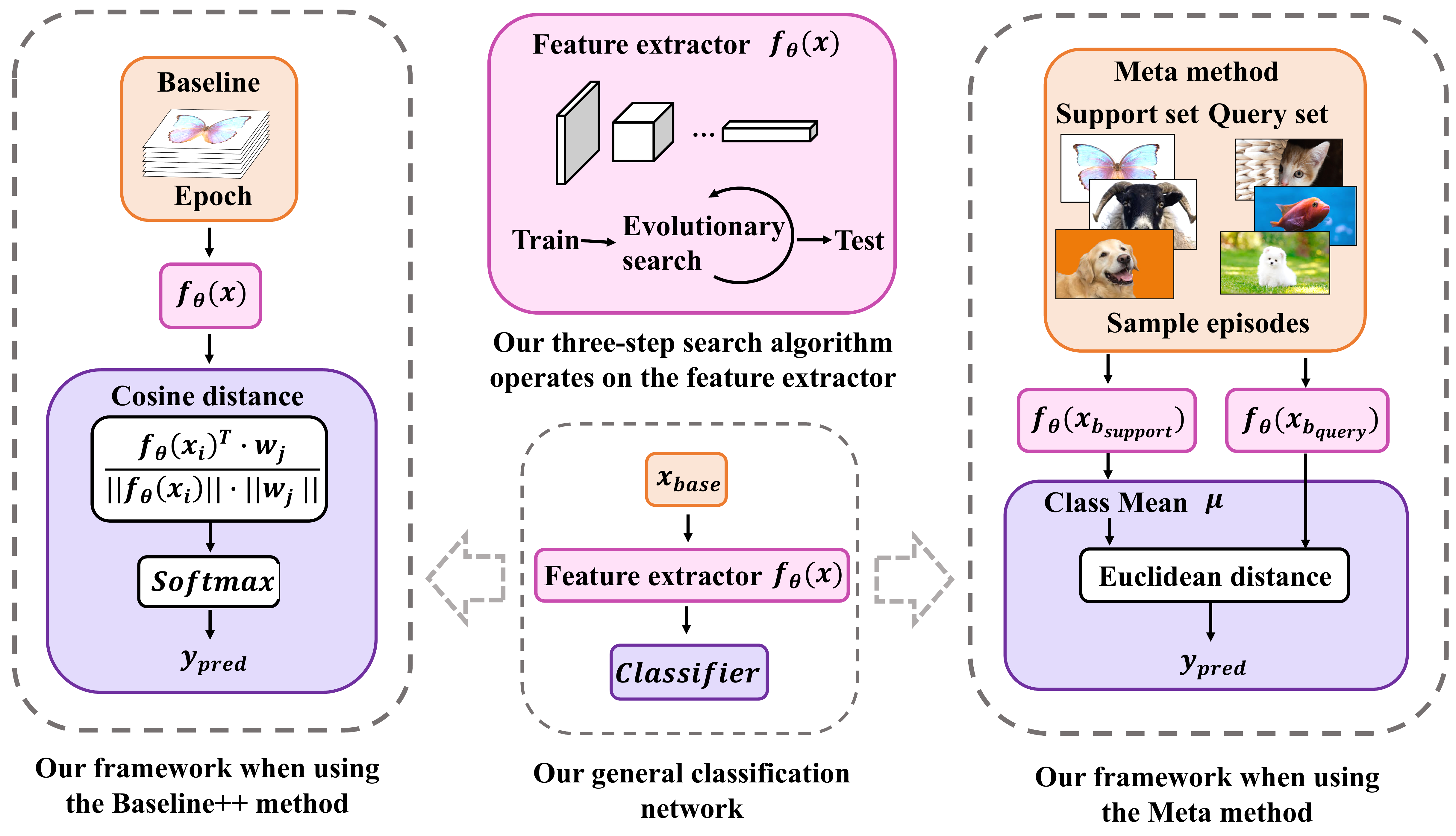}
    \vspace{-0.12in}
    \caption{In this figure, we show that our three-step search algorithm operates on the feature extractor $f_\theta(x)$. Our general framework can easily be incorporated into the baseline method with cosine distance, denoted as baseline++ \cite{chen2018a}, as well as the meta-learning based methods.}
    \label{fig:compatible_search}
    \vspace{-0.13in}
\end{figure*}

\section{3. Methodology}

In this section, we start by introducing the problem definition of few-shot classification, then we present our whole framework, which consists of three steps: 1) train a base model on base class samples (left sub-illustration in Figure~\ref{fig:overall_framework}), 2) apply evolutionary search to explore optimal transfer strategy based on accuracy metric (middle sub-figure, curve arrow indicates looping), and 3) transfer base model to novel class with the searched strategy through partially fine-tuning. Lastly, we elaborate how to design our search space for transferring and present our search algorithm in detail.

\subsection{3.1. Preliminary and Definition}
In the few-shot classification task, given abundant labeled images ${\bf X}_b$ in base classes ${\bf L}_b$ and a small proportion of labeled images ${\bf X}_n$ in novel classes ${\bf L}_n$, ${\bf L}_b \cap {\bf L}_n=\emptyset$. Our goal is to train models for recognizing novel classes with the labeled large amount of base data and limited novel data. Considering an $N$-way $K$-shot few-shot task, where the support set on novel class has $N$ classes with $K$ labeled images and the query set contains the same $N$ classes with $Q$ unlabeled images in each class, the few-shot classification algorithms are required to learn classifiers for recognizing the $N\times Q$ images in the query set of $N$ classes.

Our objective of P-Transfer aims to discover the best transfer learning scheme $V_{lr}^*$, such that, the network achieves maximal accuracy when fine-tuning under that scheme:
\begin{align}
    V_{lr}^* = \arg \max \mathcal{A}cc(W, V_{lr}),
\end{align}
where $V_{lr} = [v_1,v_2,...,v_L]$ defines the layer-wise learning rate for fine-tuning the feature extractor, $W$ is the network's parameters and $L$ is the total number of layers.

\subsection{3.2. Framework}
As shown in Figure~\ref{fig:overall_framework}, our method consists of three steps: base class pre-training, evolutionary search, and partially transfer based on the searched strategy.

\noindent{\textbf{Step 1: Base class pre-training.}} Base class pre-training is the fundamental step of our whole pipeline. As shown in Figure~\ref{fig:overall_framework} (a), for the simple baseline, we follow the common practice to train the model from scratch by minimizing a standard cross-entropy objective with the training samples in base classes. For the meta-learning pipeline, the meta-pretraining also follows the conventional strategy that a meta-learning classifier is conditioned on the base support set. More
specifically, in the meta-pretraining stage, support set and query set on the base class are first sampled randomly from $N$ classes, and then train the parameters to minimize the $N$-way prediction loss.

\noindent{\textbf{Step 2: Evolutionary search.}} The second step is to perform evolutionary search with different fine-tuning strategies to determine which layers will be fixed and others will be fine-tuned in the representation transfer stage. We also consider the above two circumstances: simple baseline through pre-training + fine-tuning, and meta-based methods. In these two scenarios the evolutionary searching operations are slightly different, as shown in Figure~\ref{fig:overall_framework} (b) and Figure~\ref{fig:compatible_search}. Generally, our method searches the optimal strategy for transferring from base classes to novel classes through fixing or re-activating some particular layers that can help novel classes. As this is the core of our framework, we will elaborate in the following sections individually (Section 3.4 and 3.5).

\noindent{\textbf{Step 3: Partially transfer via searched strategy.}} As shown in Figure~\ref{fig:overall_framework} (c), the final step is to apply our searched transfer strategy to the novel classes. Different from the simple baseline that fixes backbone and fine-tunes the last linear layer only, or meta-learning methods that use the base network as a feature extractor for the meta-testing, we will partially fine-tune our base network on the novel support set based on the searched strategies for both types of methods. This is also the core component to achieve significant improvement.

\subsection{3.3. Search Space}

Our search space is related to the model architecture we utilize for the few-shot classification. Generally, it contains the layer-level selection (fine-tuning or freezing) and learning rate assignment for fine-tuning. The search space can be formulated as $m^K$, where $m$ is the number of choices for learning rate values and $K$ is the number of layers in networks. For example, if we choose $\text{learning rate}\in\{0, 0.01, 0.1, 1.0\}$ as a learning rate zoo (``learning rate $=0$'' indicates we freeze this layer during fine-tuning), i.e., $m$ = 4. For Conv6 structure, the search space includes $4^6$ possible transfer strategies. Our searching method can automatically match the optimal choice for each layer from the learning rate zoo during fine-tuning. A brief comparison of the search space is described in Table~\ref{tab:search_space-table}. It increases sharply if we choose deeper networks.

\begin{table}[h]
\vspace{-0.1in}
\centering
\caption{Search Space of P-Transfer.}
\label{tab:search_space-table}
\vspace{-0.09in}
\resizebox{.37\textwidth}{!}{%
\begin{tabular}{ccccc}
 \toprule[1.1pt]
  \bf Network    &  Conv6 & ResNet-12  &  ResNet-$K$    \\ \midrule
 \bf Complexity     &  $m^6$   &   $m^{12}$ &   $m^K$   \\
\bottomrule[1.1pt]
\end{tabular}
}
\vspace{-0.12in}
\end{table}

\subsection{3.4. Search Algorithm} \label{search}

Our searching step is following the evolutionary algorithm. Evolutionary algorithms, a.k.a genetic algorithms, base on the natural evolution of creature species. It contains reproduction, crossover, and mutation stages. Here in our scenario, first a population of strategies is embedded to vectors $\mathcal V$ and initialized randomly. Each individual $v$ consists of its strategy for fine-tuning. After initialization, we start to evaluate each individual strategy $v$ to obtain its accuracy on the validation set. Among these evaluated strategies we select the top $K$ as parents to produce posterity strategies. The next generation strategies are made by mutation and crossover stages. By repeating this process in iterations, we can find a best fine-tuning strategy with the best validation performance. The detailed search pipeline is presented in Algorithm~\ref{alg:1} and the hyper-parameters for this algorithm are introduced in Section~4.

In this work we conduct the evolutionary search in transfer learning for few-shot classification. We target at fine-tuning with diverse learning rates to explore suitable transfer patterns in terms of knowledge with a simple and effective strategy design. At each layer, the learning rate is selected from a pre-defined zoo with all possible choices.

\subsection{3.5. Incorporating into Few-Shot Classification Frameworks} \label{applying}
As in Figure~\ref{fig:compatible_search}, we introduce how to incorporate our search algorithm into existing few-shot classification frameworks. We choose the non-meta baseline++~\cite{chen2018a} and meta ProtoNet~\cite{snell2017prototypical} as examples.

\noindent{\textbf{Upon simple baseline++.}} Baseline++ aims to explicitly reduce intra-class variation among features by applying cosine distances between the feature and weight vector in the training and fine-tuning stages. As shown in Figure~\ref{fig:compatible_search} (left sub-figure), we follow the design of distance-based classifier in searching but adjust the backbone feature extractor $f_\theta(x)$ through exploring different learning rates for different layers during fine-tuning. Intuitively, the learned backbone and distance-based classifier from our searching method are more harmonious and powerful than freezing backbone network and only fine-tuning weight vectors for few-shot classification, since our whole model is tuned end-to-end. 

\vspace{-0.15in}
\begin{algorithm}
\caption{Evolutionary algorithm for searching the best fine-tuning configuration.}
\label{alg:1}
\textbf{Input:} Trained feature extractor: $\mathcal{N}$, layer index in a network: $l$, the meta-validation loss: $\mathcal{L}$, number of {\em Random} sampling operation: $R$, number of {\em Mutation}: $M$, number of {\em Crossover}: $C$, max number of {\em Iterations}: $I$. \\
\textbf{Output}: Optimized fine-tuning configuration $v^*$\\
\begin{algorithmic}
\STATE $ \textbf{define}$ miniEval($v$):
\STATE $\ \ \ \ $ $\mathcal{N}_i$ = Load( $\mathcal{N}$) \ \ \ \# Inherit the weights from the feature extractor trained on the base dataset.
\STATE $\ \ \ \ $ Set $grad_{l} = 0$ $  \textbf{if}  $ $v_i$[$l$] = 0  \ \ \ \# Set the gradient to 0 according to the scheme vector.
\STATE $\ \ \ \ $ \{$v_i$, accuracy\} = miniFinetune($\mathcal{N}_i$) \ \ \# Fine-tune the targeting layers.
\STATE $\ \ \ \ $ $\textbf{return}$ \{$v_i$, accuracy\}
\STATE 
\vspace{-0.1in}
\FOR{$i = 0:R$}
\STATE $v_i$ = RandomChoice([0, m], $\mathcal{L}$) \ \ \ \# Randomly sample fine-tuning schemes, i.e., chose $lr$ for each layer.
\STATE \{$v_i$, accuracy\} = miniEval($v_i$) \ \ \ \# Evaluate accuracy on the validation dataset.
\ENDFOR
\STATE $v_{topK}$ =  Top$K$(\{$V$, accuracy\}) \ \ \ \# Initialize the population with Top$K$ vectors. $V$ is the set of $\{v_i\}$.
\FOR{$j = 0:I$}
\FOR{$i = 0:M$}
\STATE $v_i$ = Mutation($v_{topK}$, $\mathcal{L}$) \ \ \ \# Generate the off-spring fine-tuning vectors based on the top ones.
\STATE \{$v_i$, accuracy\} = miniEval($v_i$)\ \ \ \# Evaluate off-springs' accuracy on the validation dataset.
\ENDFOR
\FOR{$i = 0:C$}
\STATE $v_i$ = Crossover(\{$v_{topK_1}, v_{topK_2}$\}, $\mathcal{L}$) \ \ \# Generate the {\em crossover} vectors between two parents.
\STATE \{$v_i$, accuracy\} = miniEval($v_i$)\ \ \ \# Evaluate off-springs' accuracy on the validation dataset.
\ENDFOR
\STATE $v_{topK}$ =  Top$K$(\{$V$, accuracy\}) \ \ \ \# Update the population by choosing the Top$K$ vectors.
\ENDFOR
\STATE $v^*$, $acc^*$= Top1(\{$V$, accuracy\}) \ \ \ \# Select the best scheme vector with highest validation accuracy.
\STATE \textbf{return} $v^*$;
\end{algorithmic}
\end{algorithm}
\vspace{-0.15in}

\noindent{\textbf{Upon meta-learning based methods.}} As shown in Figure~\ref{fig:compatible_search} (right sub-figure), we describe the formulation of how to apply our searching method to meta-learning method for few-shot classification. In the meta-training stage, the algorithm first randomly chooses $N$ classes, and samples small base support set $x_{b{(s)}}$ and a base query set $x_{b{(q)}}$ from samples within these classes. The objective is to learn a classification model $M$ that minimizes $N$-way prediction loss of the samples in the query set $Q_b$. Here, the classifier $M$ is conditioned on the provided support set $x_b$. Similar to baseline++, we train the classification model $M$ by fine-tuning the backbone network and classifier simultaneously, to discover the optimal fine-tuning strategy.
As the predictions from a meta-based classifier are conditioned on the given support set, the meta-learning method can learn to learn from limited labeled data through a collection of episodes.

\begin{table*}[t]
\vspace{-1em}
\begin{center}
\caption{We validate on the non-meta method with Conv6 structure. We report the mean of 600 randomly generated episodes and the 95\% confidence intervals. We compare the original learning algorithm (i.e., fine-tuning the fully-connected layer only and referring as ``Fixed'' in the table) with fine-tuning the human-defined last convolutional layer (i.e., ``Manual'' in the table) and fine-tuning the layers based on the evolutionary-searched scheme (i.e., ``Searched'' in the table).}
\label{table:mcv6_flops}
\vspace{-1em}
\resizebox{.52\textwidth}{!}{
\begin{tabular}{ccccccccc}
\noalign{\vspace{1.5pt}}
\hline
\hline
\noalign{\vspace{1.5pt}}
& \multicolumn{2}{c}{CUB} & \multicolumn{2}{c}{{\em mini}-ImageNet} \\
\noalign{\vspace{1.5pt}}
\cline{2-5}
\noalign{\vspace{1.5pt}}
& 1-shot & 5-shot & 1-shot & 5-shot \\
\noalign{\vspace{1.5pt}}
\hline
\noalign{\vspace{1.5pt}}
Fixed & \textbf{66.75}$\pm$\textbf{0.91} & 79.48$\pm$0.63 & 52.52$\pm$0.79 & 71.16$\pm$0.66\\
Manual & 64.27$\pm$0.97 & 79.26$\pm$0.61 & 53.22$\pm$0.77 & 71.43$\pm$0.64\\
Searched & 65.82$\pm$0.97 & \textbf{80.48}$\pm$\textbf{0.60} & \textbf{53.55}$\pm$\textbf{0.80} & \textbf{71.45}$\pm$\textbf{0.68}\\
\noalign{\vspace{1.5pt}}
\hline
\hline
\end{tabular}
}
\end{center}
\vspace{-0.20in}

\begin{center}
\caption{Few-shot classification results on {\em mini}-ImageNet and CUB datasets with ResNet-12 structure. We apply our proposed algorithm to Baseline++ (non-meta) few-shot method as well as the meta-learning method (ProtoNet). The results show that in most cases, the proposed algorithm can discover a better knowledge transferring scheme than the original scheme.}
\vspace{-0.1em}
\label{table:res12_flops}
\resizebox{.67\textwidth}{!}{
\begin{tabular}{ccccccccc}
\hline
\hline
\noalign{\vspace{1.5pt}}
& \multicolumn{2}{c}{CUB} & \multicolumn{2}{c}{{\em mini}-ImageNet} \\
\noalign{\vspace{1.5pt}}
\cline{2-5}
\noalign{\vspace{1.5pt}}
& 1-shot & 5-shot & 1-shot & 5-shot \\
\noalign{\vspace{1.5pt}}
\hline
\noalign{\vspace{1.5pt}}
\multicolumn{2}{l}{\textbf{Few-shot learning method: Baseline++}}\\
\noalign{\vspace{1.5pt}}
\hline
\noalign{\vspace{1.5pt}}
Fixed & 70.72$\pm$0.88 & 85.59$\pm$0.54& 59.35$\pm$0.82 & 77.51$\pm$0.59 \\
Searched & \textbf{73.88}$\pm$\textbf{0.87} & \textbf{87.81}$\pm$\textbf{0.48} & \textbf{64.21}$\pm$\textbf{0.77} & \textbf{80.38}$\pm$\textbf{0.59} \\
\noalign{\vspace{1.5pt}}
\hline
\noalign{\vspace{1.5pt}}
\multicolumn{2}{l}{\textbf{Few-shot learning method: ProtoNet}}\\
\noalign{\vspace{1.5pt}}
\hline
\noalign{\vspace{1.5pt}}
Fixed & \textbf{73.82}$\pm$\textbf{0.92}& 87.28$\pm$0.48& 53.88$\pm$0.81& 74.87$\pm$0.67\\
Searched & 73.16$\pm$0.92& \textbf{88.32}$\pm$\textbf{0.46}& \textbf{54.36}$\pm$\textbf{0.81} & \textbf{76.59}$\pm$\textbf{0.64}\\
\hline
\hline
\end{tabular}
}
\end{center}
\vspace{-0.1in}
\end{table*}

\begin{figure*}[h]
    \vspace{-0.15in}
    \centering
    \includegraphics[width=0.9\linewidth]{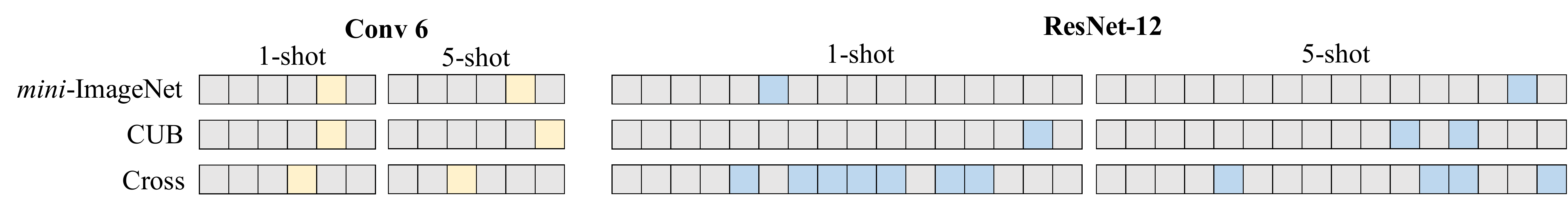}
    \vspace{-0.2in}
    \caption{We visualize the fine-tuning scheme discovered by our evolutionary algorithm. The grey boxes denote layers without fine-tuning, the colored boxes denote layers that require fine-tuning. Different scenarios have different searched scheme. For example, in the cross-domain transfer-learning, more layers need to be fine-tuned to adapt the knowledge for the target domain.}
    \label{fig:compatible_searched}
    \vspace{-0.18in}
\end{figure*}

\section{4. Experiments}
\textbf{Dataset.}
We verify our method for few-shot learning on both \textit{mini}-ImageNet dataset and CUB200-2012 dataset. \textit{mini}-ImageNet dataset is a commonly used dataset for few-shot classification. It consists of 100 classes from ImageNet dataset~\cite{deng2009imagenet}, and 600 images for each class. We follow~\cite{ravi2016optimization} to split the data into 64 base classes, 16 validation classes and 20 novel classes. CUB200-2011 contains 200 classes of birds~\cite{wah2011caltech}. Follow~\cite{hilliard2018few}, we split the data into 100 base classes, 50 validation classes and 50 novel classes. 
We validate the effectiveness of our method for generic classification on \textit{mini}-ImageNet, and for fine-grained classification on CUB, as well as for cross-domain adaptation through transferring knowledge learned from \textit{mini}-ImageNet to CUB.

\begin{table*}[t]
\vspace{-0.2in}
\begin{center}
\caption{In this table, we further evaluate our method on the cross-domain few-shot learning tasks, i.e., transferring the knowledge from {\em mini}-ImageNet to CUB. We conduct experiments on both meta and non-meta methods. We find that when there exist domain difference, fine-tuning more layers are required. Moreover, we further discovered using GroupNorm can also bridge the distribution difference between training and testing, which outperforms the results of using BatchNorm.}
\label{table:gn_flops}
\vspace{-0.1in}
\resizebox{.75\textwidth}{!}{
\begin{tabular}{ccccccccc}
\hline
\hline
\noalign{\vspace{1.5pt}}
& & \multicolumn{2}{c}{1-shot} & \multicolumn{2}{c}{5-shot} \\
\noalign{\vspace{1.5pt}}
\cline{3-6}
\noalign{\vspace{1.5pt}}
& & BatchNorm & GroupNorm & BatchNorm & GroupNorm \\
\noalign{\vspace{1.5pt}}
\hline
\noalign{\vspace{1.5pt}}
\multicolumn{2}{l}{\textbf{Few-shot learning method: Baseline++}}\\
\noalign{\vspace{1.5pt}}
\hline
\noalign{\vspace{1.5pt}}
\multirow{2}{*}{Conv6} & Fixed & 40.77$\pm$0.70 & 44.80$\pm$0.78 & 58.15$\pm$0.72 & 64.15$\pm$0.74 \\
& Searched & \textbf{41.69}$\pm$\textbf{0.72} & \textbf{45.34}$\pm$\textbf{0.76} & \textbf{61.53}$\pm$\textbf{0.70} & \textbf{67.42}$\pm$\textbf{0.72} \\
\noalign{\vspace{1.5pt}}
\hline
\noalign{\vspace{1.5pt}}
\multirow{2}{*}{ResNet-12} & Fixed & 43.14$\pm$0.72 & 43.52$\pm$0.71 & 63.25$\pm$0.70 & 64.02$\pm$0.71\\
& Searched & \textbf{43.77}$\pm$\textbf{0.74} & \textbf{45.86}$\pm$\textbf{0.72} & \textbf{68.30}$\pm$\textbf{0.73} & \textbf{74.22}$\pm$\textbf{0.66} \\
\noalign{\vspace{1.5pt}}
\hline
\noalign{\vspace{1.5pt}}
\multicolumn{2}{l}{\textbf{Few-shot learning method: ProtoNet}}\\
\noalign{\vspace{1.5pt}}
\hline
\noalign{\vspace{1.5pt}}
\multirow{2}{*}{Conv6} & Fixed & 36.34$\pm$0.71 & \textbf{38.74}$\pm$\textbf{0.77} & 	\textbf{55.38}$\pm$\textbf{0.71} & 59.30$\pm$0.73 \\
& Searched & \textbf{36.36}$\pm$\textbf{0.73} & 38.36$\pm$0.69 & 54.06$\pm$0.73 & \textbf{62.33}$\pm$\textbf{0.75} \\
\noalign{\vspace{1.5pt}}
\hline
\noalign{\vspace{1.5pt}}
\multirow{2}{*}{ResNet-12} & Fixed & 39.14$\pm$0.73 & 44.88$\pm$0.78 & 60.11$\pm$0.73 & 67.18$\pm$0.74 \\
& Searched &\textbf{39.38}$\pm$\textbf{0.72} & \textbf{45.29}$\pm$\textbf{0.77} & \textbf{60.24}$\pm$\textbf{0.73} & \textbf{68.42}$\pm$\textbf{0.75}\\
\noalign{\vspace{1.5pt}}
\hline
\noalign{\vspace{1.5pt}}
\hline
\end{tabular}
}
\end{center}
\vspace{-0.15in}
\end{table*}

\begin{table*}[t]
\begin{center}
\caption{Comparison with the state-of-the-art results on {\em mini}-ImageNet dataset. * denotes the results re-implemented by us. $^{\dagger}$ indicates the results are from more training techniques like DropBlock and label smoothing.}
\vspace{-0.06in}
\label{table:final_flops_results}
\vspace{-0.05in}
\resizebox{.8\textwidth}{!}{
\begin{tabular}{llcccccc}
\hline
\hline
\noalign{\vspace{1.5pt}}
Method & Backbone & 1-shot & 5-shot \\
\noalign{\vspace{1.5pt}}
\hline
\noalign{\vspace{1.5pt}}
MatchingNet~\cite{vinyals2016matching} & Conv4 & 43.56$\pm$0.84 &  55.31$\pm$0.73 \\
{\bf MatchingNet*}~\cite{vinyals2016matching} & ResNet-12 & \bf 54.76$\pm$\bf 0.82 & \bf 70.01$\pm$\bf 0.70 \\
ProtoNet~\cite{snell2017prototypical} & Conv4 & 48.70$\pm$1.84 & 63.11$\pm$0.92 \\
{\bf ProtoNet*}~\cite{snell2017prototypical} & ResNet-12 & \bf 53.88$\pm$ \bf 0.81 & \bf 74.87$\pm$ \bf 0.67 \\ \hline
Parameters from Activations~\cite{qiao2018few} & WRN-28-10 &  59.60$\pm$0.41 & 73.74$\pm$0.19 \\
Closer Look~\cite{chen2018a} & ResNet-18 & 51.87$\pm$0.77 & 75.68$\pm$0.63 \\
SNAIL~\cite{mishra2017snail} & ResNet-12 &  55.71$\pm$0.99 & 68.88$\pm$0.92 \\
AdaResNet~\cite{munkhdalai2017rapid} & ResNet-12 & 56.88$\pm$0.62 & 71.94$\pm$0.57 & \\
TADAM~\cite{oreshkin2018tadam} & ResNet-12 &  58.50$\pm$0.30 & 76.70$\pm$0.30 \\
MetaOptNet~\cite{lee2019meta} & ResNet-12 &  60.33$\pm$0.61 & 76.61$\pm$0.46 \\
MetaOptNet$^{\dagger}$~\cite{lee2019meta} & ResNet-12 &  62.64$\pm$0.61 & 78.63$\pm$0.46 \\
Meta-Baseline~\cite{chen2020new} & ResNet-12 &  63.17$\pm$ 0.23  & 79.26$\pm$ 0.17 \\
\noalign{\vspace{1.5pt}}
\hline
\noalign{\vspace{1.5pt}}
P-Transfer (ours) & ResNet-12 & \textbf{64.21}$\pm$\textbf{0.77} & \textbf{80.38}$\pm$\textbf{0.59}\\
\hline
\hline
\end{tabular}
}
\end{center}
\vspace{-0.15in}
\end{table*}

\noindent{\textbf{Implementation.}} \label{exp}
For meta methods, we sample episodes with 5 classes from the target dataset. Then for each class, we sample $k$ instances as the support set and 15 instances as the query set for a $k$-shot task. In training, we train 60,000 episodes for 1-shot and 40,000 episodes for 5-shot tasks on the base dataset. In search, we sample 20 episodes from the validation dataset. We fine-tune the network on the support set for 100 iterations and evaluate the network on the query set. In evaluation, We fine-tune layers following the searched configuration on the support set and evaluate on the query set with episodes sampled from novel dataset. We ran 600 episodes and report the average accuracy and the 95\% confidence intervals. The non-meta method differs only in the training stage, where we train the feature extractor for 400 epochs with a batchsize of 16 on the base dataset.

We adopt Adam optimizer with learning rate of 1e-3 for training. In fine-tuning, we use SGD with 0.01 learning rate for fully-connected layer and other searched learning rates for the corresponding layers. 
We use standard data augmentation including random crop, horizontal flip and color jitter. For Algorithm~\ref{alg:1}, we set population size  $P=20$, max iterations $I=20$, and number of {\em random sampling} ($R$), {\em mutation} ($M$) and {\em crossover} ($C$) to $50$.

\subsection{4.1. Ablation Study}
\noindent{\textbf{Comparison to fixed and manually designed fine-tuning.}}
We first compare our proposed method with fixed and manually designed fine-tuning schemes using Conv6 and ResNet-12 structures on CUB and {\em mini}-ImageNet. The reason that we compare with fine-tuning the last convolutional layer as generally, the last layer is more domain-specific. Thus, in manually designed fine-tuning scheme, researchers usually fine-tune the last convolutional layer as a solution. Our results are shown in Table~\ref{table:mcv6_flops} (Conv6) and~\ref{table:res12_flops} (ResNet-12), where in general, our evolutionary strategy achieves better accuracy than fixing backbone and human-defined strategy. As a baseline, we obtain 40.84$\pm$0.8\% (1-shot) and 50.95$\pm$0.9\% (5-shot) when fine-tuning all the layers on {\em mini}-ImageNet.

\noindent{\textbf{Comparison to different normalization layers in cross-domain setting.}} As fine-tuning backbone networks will significantly be affected by the size of batchsize, while for the few-shot classification scenario, we do not have enough samples to increase the batchsize, and also the conventional fixing backbone solutions do not encounter this problem. Thus, here we further explore whether a better batch norm technique can deliver further improvement. Our results are shown in Table~\ref{table:gn_flops}, in the cross-domain settings, group norm~\cite{wu2018group} can achieve much better accuracy (about 2$\sim$8\% higher) than batch norm~\cite{ioffe2015batch} since it can overcome the drawback of optimization issue from small batchsize in traditional batch norm. 
Nevertheless, for a fair comparison, we only apply group norm in this ablation study to verify our conjecture that the limited batchsize may be a restriction to fully liberate the effectiveness and potential of our partial transfer method during backbone fine-tuning. As other state-of-the-art methods used standard batch norm, in our other results we still use the same batch norm method.

\subsection{4.2. Searched Schemes and Final Results}
To better understand our partial transfer method, we visualize the searched schemes in Figure~\ref{fig:compatible_searched}. We observe two interesting phenomena which are in line with the intuition:
(1) Deeper networks will always have more layers that require to be fine-tuned for few-shot learning;
(2) When the domain difference between base and novel data is increased (in the cross-domain scenario), more layers are required to be fine-tuned.

Our final results are shown in Table~\ref{table:final_flops_results}, we can see that our partial transfer method can consistently outperform other state-of-the-art on both 1 and 5 shots settings. Even without additional training techniques like DropBlock~\cite{ghiasi2018dropblock} and label smoothing~\cite{szegedy2016rethinking}, our method still obtains a significant improvement, as our flexible transfer/fine-tuning can benefit from few support samples to adjust the backbone parameters.

\subsection{4.3. Extension to Traditional Transfer Learning}
We further explore the traditional transfer learning from ImageNet~\cite{deng2009imagenet} to CUB200-2012 with the Inception V3 network~\cite{szegedy2016rethinking}. We use  SGD optimizer with initial learning rate being 0.01 and linearly decay to 0. In transferring, we observe that, the weights learned from our method i.e., re-initializing and fine-tuning a few layers for partial transfer achieves higher accuracy than inherit all the weights and do fine-tuning, as shown in Table~\ref{I_to_cub}.

\begin{table}[h]
\begin{center}
\caption{The comparison between inheriting all to partially reinitialize weights in transfer learning.}
\label{I_to_cub}
\vspace{-0.1in}
\begin{tabular}{cccccccc}
\toprule[1.0pt]
\noalign{\vspace{1.5pt}}
& \bf Baseline &\bf Partial transfer\\
Top-1 Accuracy &  82.9\% & \bf 83.8\% \\
\noalign{\vspace{1.5pt}}
\bottomrule[1.0pt]
\end{tabular}
\end{center}
\vspace{-0.18in}
\end{table}

\section{5. Discussions}

\subsection{5.1. The Essential Role of Transferring in Few-shot}
In the setting of few-shot learning, the pre-trained feature extractor is required to provide proper transferability from base to novel class in the meta or non-meta learning stage. Basically, the transferring learning aims to transfer the common knowledge from base objects to the novel class. However, as we stated above, there definitely has some unnecessary and even harmful information in the base class, since the novel data is few and sensitive to the feature extractor, the complete transferring strategy will not be able to avoid them, indicating that our method is a better solution for the few-shot scenario. 

\subsection{5.2. Why Is Partial Better Than All in Few-shot Learning?}
Usually the base and novel class are in the same domain, so using the pre-trained feature extractor on base data and then transferring to novel data can obtain good or moderate performance. However, as shown in Figure~\ref{fig:compatible_searched}, in the cross-domain transfer-learning, more layers need to be fine-tuned to adapt the knowledge for the target domain since the source and target domains are discrepant in content. In this circumstance, the conventional transfer learning is no longer applicable. Our proposed partial transferring with diverse learning rates on different layers is competent for this intractable situation, and intuitively, fixed transferring is generally a special case of our strategy and ours has better potential in few-shot learning.

\section{6. Conclusion}
We have introduced a partial transfer (P-Transfer) method for the few-shot classification. Our method is the first attempt to thoroughly explore the capability of transferring through searching strategies in few-shot scenario without any proxy. Our method boosts both the meta and non-meta based methods by a large margin under 1-shot or 5-shot circumstances, as our flexible transfer/fine-tuning can benefit from few support samples to adjust the backbone parameters. Intuitively, our partial transfer has larger potential for few-shot classification and even for the traditional transfer learning. We hope our method can inspire more methods along this direction. In the future, we will perform more analyses about how partial transfer helps the few-shot problems. We will apply our method on other few-shot tasks like detection, segmentation to explore the upper limit of our proposed transfer method.

\section*{Potential Ethical Impact}
This work has the following potential positive and negative impacts in the society. As machine learning algorithm has been highly studied in data-intensive applications, such as the large-scale ImageNet classification, but is often hampered when the training data is small. This work tackles this problem through using prior knowledge from other sufficient labeled data, and can rapidly generalize to new data containing only a few samples with annotated information. It can help researchers or industry to build systems in the area that data is costly to collect, such as medical images or wild animals. However, we should be cautious of the result of failure of the system which could cause unreliable conclusion, such as the misclassified disease for medical images and further was somewhat misleading to the doctors.

{\small
\bibliographystyle{ieee_fullname}
\bibliography{my}

\begin{thebibliography}{42}
\providecommand{\natexlab}[1]{#1}
\providecommand{\url}[1]{\texttt{#1}}
\providecommand{\urlprefix}{URL }
\expandafter\ifx\csname urlstyle\endcsname\relax
  \providecommand{\doi}[1]{doi:\discretionary{}{}{}#1}\else
  \providecommand{\doi}{doi:\discretionary{}{}{}\begingroup
  \urlstyle{rm}\Url}\fi

\bibitem[{Andrychowicz et~al.(2016)Andrychowicz, Denil, Gomez, Hoffman, Pfau,
  Schaul, Shillingford, and De~Freitas}]{andrychowicz2016learning}
Andrychowicz, M.; Denil, M.; Gomez, S.; Hoffman, M.~W.; Pfau, D.; Schaul, T.;
  Shillingford, B.; and De~Freitas, N. 2016.
\newblock Learning to learn by gradient descent by gradient descent.
\newblock In \emph{Advances in neural information processing systems},
  3981--3989.

\bibitem[{Antoniou, Edwards, and Storkey(2018)}]{antoniou2018train}
Antoniou, A.; Edwards, H.; and Storkey, A. 2018.
\newblock How to train your MAML.
\newblock \emph{arXiv preprint arXiv:1810.09502} .

\bibitem[{Chen et~al.(2019)Chen, Liu, Kira, Wang, and Huang}]{chen2018a}
Chen, W.-Y.; Liu, Y.-C.; Kira, Z.; Wang, Y.-C.~F.; and Huang, J.-B. 2019.
\newblock A Closer Look at Few-shot Classification.
\newblock In \emph{International Conference on Learning Representations}.

\bibitem[{Chen et~al.(2020)Chen, Wang, Liu, Xu, and Darrell}]{chen2020new}
Chen, Y.; Wang, X.; Liu, Z.; Xu, H.; and Darrell, T. 2020.
\newblock A new meta-baseline for few-shot learning.
\newblock \emph{arXiv preprint arXiv:2003.04390} .

\bibitem[{Deng et~al.(2009)Deng, Dong, Socher, Li, Li, and
  Fei-Fei}]{deng2009imagenet}
Deng, J.; Dong, W.; Socher, R.; Li, L.-J.; Li, K.; and Fei-Fei, L. 2009.
\newblock Imagenet: A large-scale hierarchical image database.
\newblock In \emph{2009 IEEE conference on computer vision and pattern
  recognition}, 248--255. Ieee.

\bibitem[{Elsken et~al.(2020)Elsken, Staffler, Metzen, and
  Hutter}]{elsken2020meta}
Elsken, T.; Staffler, B.; Metzen, J.~H.; and Hutter, F. 2020.
\newblock Meta-Learning of Neural Architectures for Few-Shot Learning.
\newblock In \emph{Proceedings of the IEEE/CVF Conference on Computer Vision
  and Pattern Recognition}, 12365--12375.

\bibitem[{Finn, Abbeel, and Levine(2017)}]{finn2017model}
Finn, C.; Abbeel, P.; and Levine, S. 2017.
\newblock Model-agnostic meta-learning for fast adaptation of deep networks.
\newblock In \emph{Proceedings of the 34th International Conference on Machine
  Learning-Volume 70}, 1126--1135. JMLR. org.

\bibitem[{Garcia and Bruna(2017)}]{garcia2017few}
Garcia, V.; and Bruna, J. 2017.
\newblock Few-shot learning with graph neural networks.
\newblock \emph{arXiv preprint arXiv:1711.04043} .

\bibitem[{Ghiasi, Lin, and Le(2018)}]{ghiasi2018dropblock}
Ghiasi, G.; Lin, T.-Y.; and Le, Q.~V. 2018.
\newblock Dropblock: A regularization method for convolutional networks.
\newblock In \emph{Advances in Neural Information Processing Systems},
  10727--10737.

\bibitem[{Gidaris and Komodakis(2018)}]{gidaris2018dynamic}
Gidaris, S.; and Komodakis, N. 2018.
\newblock Dynamic few-shot visual learning without forgetting.
\newblock In \emph{Proceedings of the IEEE Conference on Computer Vision and
  Pattern Recognition}, 4367--4375.

\bibitem[{He et~al.(2016)He, Zhang, Ren, and Sun}]{he2016deep}
He, K.; Zhang, X.; Ren, S.; and Sun, J. 2016.
\newblock Deep residual learning for image recognition.
\newblock In \emph{Proceedings of the IEEE conference on computer vision and
  pattern recognition}, 770--778.

\bibitem[{Hilliard et~al.(2018)Hilliard, Phillips, Howland, Yankov, Corley, and
  Hodas}]{hilliard2018few}
Hilliard, N.; Phillips, L.; Howland, S.; Yankov, A.; Corley, C.~D.; and Hodas,
  N.~O. 2018.
\newblock Few-shot learning with metric-agnostic conditional embeddings.
\newblock \emph{arXiv preprint arXiv:1802.04376} .

\bibitem[{Huang et~al.(2017)Huang, Liu, Van Der~Maaten, and
  Weinberger}]{huang2017densely}
Huang, G.; Liu, Z.; Van Der~Maaten, L.; and Weinberger, K.~Q. 2017.
\newblock Densely connected convolutional networks.
\newblock In \emph{Proceedings of the IEEE conference on computer vision and
  pattern recognition}, 4700--4708.

\bibitem[{Ioffe and Szegedy(2015)}]{ioffe2015batch}
Ioffe, S.; and Szegedy, C. 2015.
\newblock Batch normalization: Accelerating deep network training by reducing
  internal covariate shift.
\newblock \emph{arXiv preprint arXiv:1502.03167} .

\bibitem[{Kim, Kim, and Kim(2020)}]{kimmodel}
Kim, J.; Kim, H.; and Kim, G. 2020.
\newblock Model-Agnostic Boundary-Adversarial Sampling for Test-Time
  Generalization in Few-Shot learning.
\newblock In \emph{ECCV}.

\bibitem[{Krizhevsky, Sutskever, and Hinton(2012)}]{krizhevsky2012imagenet}
Krizhevsky, A.; Sutskever, I.; and Hinton, G.~E. 2012.
\newblock Imagenet classification with deep convolutional neural networks.
\newblock In \emph{Advances in neural information processing systems},
  1097--1105.

\bibitem[{Lee et~al.(2019)Lee, Maji, Ravichandran, and Soatto}]{lee2019meta}
Lee, K.; Maji, S.; Ravichandran, A.; and Soatto, S. 2019.
\newblock Meta-learning with differentiable convex optimization.
\newblock In \emph{Proceedings of the IEEE Conference on Computer Vision and
  Pattern Recognition}, 10657--10665.

\bibitem[{Li et~al.(2017)Li, Zhou, Chen, and Li}]{li2017meta}
Li, Z.; Zhou, F.; Chen, F.; and Li, H. 2017.
\newblock Meta-sgd: Learning to learn quickly for few-shot learning.
\newblock \emph{arXiv preprint arXiv:1707.09835} .

\bibitem[{Liu et~al.(2017)Liu, Simonyan, Vinyals, Fernando, and
  Kavukcuoglu}]{liu2017hierarchical}
Liu, H.; Simonyan, K.; Vinyals, O.; Fernando, C.; and Kavukcuoglu, K. 2017.
\newblock Hierarchical representations for efficient architecture search.
\newblock \emph{arXiv preprint arXiv:1711.00436} .

\bibitem[{Miikkulainen et~al.(2019)Miikkulainen, Liang, Meyerson, Rawal, Fink,
  Francon, Raju, Shahrzad, Navruzyan, Duffy et~al.}]{miikkulainen2019evolving}
Miikkulainen, R.; Liang, J.; Meyerson, E.; Rawal, A.; Fink, D.; Francon, O.;
  Raju, B.; Shahrzad, H.; Navruzyan, A.; Duffy, N.; et~al. 2019.
\newblock Evolving deep neural networks.
\newblock In \emph{Artificial Intelligence in the Age of Neural Networks and
  Brain Computing}, 293--312. Elsevier.

\bibitem[{Mishra et~al.(2017)Mishra, Rohaninejad, Chen, and
  Abbeel}]{mishra2017snail}
Mishra, N.; Rohaninejad, M.; Chen, X.; and Abbeel, P. 2017.
\newblock A simple neural attentive meta-learner.
\newblock \emph{arXiv preprint arXiv:1707.03141} .

\bibitem[{Munkhdalai and Yu(2017)}]{munkhdalai2017meta}
Munkhdalai, T.; and Yu, H. 2017.
\newblock Meta networks.
\newblock In \emph{Proceedings of the 34th International Conference on Machine
  Learning-Volume 70}, 2554--2563. JMLR. org.

\bibitem[{Munkhdalai et~al.(2017)Munkhdalai, Yuan, Mehri, and
  Trischler}]{munkhdalai2017rapid}
Munkhdalai, T.; Yuan, X.; Mehri, S.; and Trischler, A. 2017.
\newblock Rapid adaptation with conditionally shifted neurons.
\newblock \emph{arXiv preprint arXiv:1712.09926} .

\bibitem[{Oreshkin, L{\'o}pez, and Lacoste(2018)}]{oreshkin2018tadam}
Oreshkin, B.; L{\'o}pez, P.~R.; and Lacoste, A. 2018.
\newblock Tadam: Task dependent adaptive metric for improved few-shot learning.
\newblock In \emph{Advances in Neural Information Processing Systems},
  721--731.

\bibitem[{Qi, Brown, and Lowe(2018)}]{qi2018low}
Qi, H.; Brown, M.; and Lowe, D.~G. 2018.
\newblock Low-shot learning with imprinted weights.
\newblock In \emph{Proceedings of the IEEE conference on computer vision and
  pattern recognition}, 5822--5830.

\bibitem[{Qiao et~al.(2018)Qiao, Liu, Shen, and Yuille}]{qiao2018few}
Qiao, S.; Liu, C.; Shen, W.; and Yuille, A.~L. 2018.
\newblock Few-shot image recognition by predicting parameters from activations.
\newblock In \emph{Proceedings of the IEEE Conference on Computer Vision and
  Pattern Recognition}, 7229--7238.

\bibitem[{Ravi and Larochelle(2016)}]{ravi2016optimization}
Ravi, S.; and Larochelle, H. 2016.
\newblock Optimization as a model for few-shot learning .

\bibitem[{Real et~al.(2019)Real, Aggarwal, Huang, and Le}]{real2019regularized}
Real, E.; Aggarwal, A.; Huang, Y.; and Le, Q.~V. 2019.
\newblock Regularized evolution for image classifier architecture search.
\newblock In \emph{Proceedings of the aaai conference on artificial
  intelligence}, volume~33, 4780--4789.

\bibitem[{Real et~al.(2017)Real, Moore, Selle, Saxena, Suematsu, Tan, Le, and
  Kurakin}]{real2017large}
Real, E.; Moore, S.; Selle, A.; Saxena, S.; Suematsu, Y.~L.; Tan, J.; Le,
  Q.~V.; and Kurakin, A. 2017.
\newblock Large-scale evolution of image classifiers.
\newblock In \emph{Proceedings of the 34th International Conference on Machine
  Learning-Volume 70}, 2902--2911. JMLR. org.

\bibitem[{Simonyan and Zisserman(2014)}]{simonyan2014very}
Simonyan, K.; and Zisserman, A. 2014.
\newblock Very deep convolutional networks for large-scale image recognition.
\newblock \emph{arXiv preprint arXiv:1409.1556} .

\bibitem[{Snell, Swersky, and Zemel(2017)}]{snell2017prototypical}
Snell, J.; Swersky, K.; and Zemel, R. 2017.
\newblock Prototypical networks for few-shot learning.
\newblock In \emph{Advances in neural information processing systems},
  4077--4087.

\bibitem[{Sung et~al.(2018)Sung, Yang, Zhang, Xiang, Torr, and
  Hospedales}]{sung2018learning}
Sung, F.; Yang, Y.; Zhang, L.; Xiang, T.; Torr, P.~H.; and Hospedales, T.~M.
  2018.
\newblock Learning to compare: Relation network for few-shot learning.
\newblock In \emph{Proceedings of the IEEE Conference on Computer Vision and
  Pattern Recognition}, 1199--1208.

\bibitem[{Szegedy et~al.(2015)Szegedy, Liu, Jia, Sermanet, Reed, Anguelov,
  Erhan, Vanhoucke, and Rabinovich}]{szegedy2015going}
Szegedy, C.; Liu, W.; Jia, Y.; Sermanet, P.; Reed, S.; Anguelov, D.; Erhan, D.;
  Vanhoucke, V.; and Rabinovich, A. 2015.
\newblock Going deeper with convolutions.
\newblock In \emph{Proceedings of the IEEE conference on computer vision and
  pattern recognition}, 1--9.

\bibitem[{Szegedy et~al.(2016)Szegedy, Vanhoucke, Ioffe, Shlens, and
  Wojna}]{szegedy2016rethinking}
Szegedy, C.; Vanhoucke, V.; Ioffe, S.; Shlens, J.; and Wojna, Z. 2016.
\newblock Rethinking the inception architecture for computer vision.
\newblock In \emph{Proceedings of the IEEE conference on computer vision and
  pattern recognition}, 2818--2826.

\bibitem[{Thrun and Pratt(2012)}]{thrun2012learning}
Thrun, S.; and Pratt, L. 2012.
\newblock \emph{Learning to learn}.
\newblock Springer Science \& Business Media.

\bibitem[{Tian et~al.(2020)Tian, Wang, Krishnan, Tenenbaum, and
  Isola}]{tian2020rethinking}
Tian, Y.; Wang, Y.; Krishnan, D.; Tenenbaum, J.~B.; and Isola, P. 2020.
\newblock Rethinking Few-Shot Image Classification: a Good Embedding Is All You
  Need?
\newblock \emph{arXiv preprint arXiv:2003.11539} .

\bibitem[{Vinyals et~al.(2016)Vinyals, Blundell, Lillicrap, Wierstra
  et~al.}]{vinyals2016matching}
Vinyals, O.; Blundell, C.; Lillicrap, T.; Wierstra, D.; et~al. 2016.
\newblock Matching networks for one shot learning.
\newblock In \emph{Advances in neural information processing systems},
  3630--3638.

\bibitem[{Wah et~al.(2011)Wah, Branson, Welinder, Perona, and
  Belongie}]{wah2011caltech}
Wah, C.; Branson, S.; Welinder, P.; Perona, P.; and Belongie, S. 2011.
\newblock The caltech-ucsd birds-200-2011 dataset .

\bibitem[{Wu and He(2018)}]{wu2018group}
Wu, Y.; and He, K. 2018.
\newblock Group normalization.
\newblock In \emph{Proceedings of the European Conference on Computer Vision
  (ECCV)}, 3--19.

\bibitem[{Xie and Yuille(2017)}]{xie2017genetic}
Xie, L.; and Yuille, A. 2017.
\newblock Genetic cnn.
\newblock In \emph{Proceedings of the IEEE international conference on computer
  vision}, 1379--1388.

\bibitem[{Ye et~al.(2020)Ye, Hu, Zhan, and Sha}]{ye2020few}
Ye, H.-J.; Hu, H.; Zhan, D.-C.; and Sha, F. 2020.
\newblock Few-shot learning via embedding adaptation with set-to-set functions.
\newblock In \emph{Proceedings of the IEEE/CVF Conference on Computer Vision
  and Pattern Recognition}, 8808--8817.

\bibitem[{Zhang et~al.(2020)Zhang, Cai, Lin, and Shen}]{zhang2020deepemd}
Zhang, C.; Cai, Y.; Lin, G.; and Shen, C. 2020.
\newblock DeepEMD: Few-Shot Image Classification with Differentiable Earth
  Mover's Distance and Structured Classifiers.
\newblock In \emph{Proceedings of the IEEE/CVF Conference on Computer Vision
  and Pattern Recognition}, 12203--12213.

\end{thebibliography}
}

\end{document}